\documentclass{article} 
\usepackage{iclr2027_conference,times}

\usepackage{amsmath,amsfonts,bm}

\def\eqref#1{equation~\ref{#1}}

\def\1{\bm{1}}

\DeclareMathAlphabet{\mathsfit}{\encodingdefault}{\sfdefault}{m}{sl}
\SetMathAlphabet{\mathsfit}{bold}{\encodingdefault}{\sfdefault}{bx}{n}

\usepackage{hyperref}
\usepackage{url}
\usepackage{booktabs}
\usepackage{graphicx}
\usepackage[table]{xcolor}
\usepackage{booktabs}
\usepackage{multirow}
\usepackage{etoolbox}
\usepackage{graphicx}
\usepackage{booktabs}
\usepackage{graphicx}
\usepackage[table]{xcolor}
\usepackage{wrapfig}
\usepackage{enumitem}
\usepackage{booktabs}
\usepackage{tabularx}

\definecolor{ragheader}{RGB}{252,230,218}
\definecolor{memoryheader}{RGB}{229,234,247}

\newcounter{obs}
\usepackage{pifont}
\newcommand{\obs}[1]{%
    \par\medskip
    \refstepcounter{obs}%
    \noindent\textbf{Obs.~\theobs.\ #1}\enspace\ignorespaces
}

\newcounter{qnum}
\newcommand{\Q}{Q\theqnum}

\makeatletter
\pretocmd{\@sect}{%
    \in@{\Q}{#8}%
    \ifin@\stepcounter{qnum}\fi
}{}{\errmessage{Q patch failed}}
\makeatother

\title{Org-Agent: Beyond Personal Assistants \\ Towards Organizational Agents}

\author{
Luyao Zhuang$^{1}$,
Yujing Zhang$^{1}$,
Zijin Hong$^{1}$,
Yilin Xiao$^{1}$\thanks{Corresponding author: Yilin Xiao.}  ,
Xiao Huang$^{1}$\\
$^{1}$The Department of Computing, The Hong Kong Polytechnic University, Hong Kong SAR\\
\texttt{\{luyao.zhuang,yu-jing.zhang\}@connect.polyu.hk}; \\
\texttt{\{zijin.hong,yilin.xiao\}@connect.polyu.hk}; \\
\texttt{xiao.huang@polyu.edu.hk}
}

\iclrfinalcopy 
\begin{document}

\maketitle

\lhead{Preprint}

\begin{abstract}
Language model agents serving organizations must coordinate requests from multiple users while using knowledge distributed across their interactions. We identify two complementary capabilities for this setting, namely cross-user interaction and decision-making, as well as cross-user memory and knowledge use. Both capabilities are governed by organizational constraints across three aspects: user identity, authority, and access permissions; the attribution and temporal validity of information; and rules for resolving conflicting requirements across users and completion requirements for joint decisions. These constraints shape what information or decisions must be obtained before an action can proceed and what conditions must be satisfied during its execution. Motivated by this, we introduce \textsc{Org-Agent}, a unified constraint-centric reasoning framework that organizes task execution in three stages. Specifically, \textsc{Org-Agent} decomposes a task into atomic subtasks and constructs a task dependency graph whose edges encode the dependencies among them. 
Building on this graph, it schedules the subtasks in dependency order through topological sorting. It then executes each subtask while accounting for the task's constraints, supported by evidence-acquisition and memory-management tools. Experiments on MUSES-Bench and GroupMemBench demonstrate the effectiveness of \textsc{Org-Agent} on both capabilities, and ablations further support the contributions of dependency modeling and tool use. 
\end{abstract}

\section{Introduction}

Language model agents~\citep{zhou2026qwen,singh2025openai,deepseekai2026deepseekv41flashpushinglimitskv} offer a path toward automating organizational workflows~\citep{xu2025theagentcompany}. These workflows often involve multiple users, each holding part of the information and authority needed to complete tasks. However, existing agent systems are largely designed and evaluated for a single user~\citep{huang2026surveyagentmemorysecond,chhikara2025mem0,yao2025taubench}. Extending these agents to organizational settings is not merely a matter of scale, but of coordinating requests across users and using knowledge distributed among them. This motivates our study of \emph{agents for organizations}, where one shared agent serves multiple users and supports the completion of their tasks.

\begin{figure}[t]
    \centering
    \includegraphics[width=\textwidth]{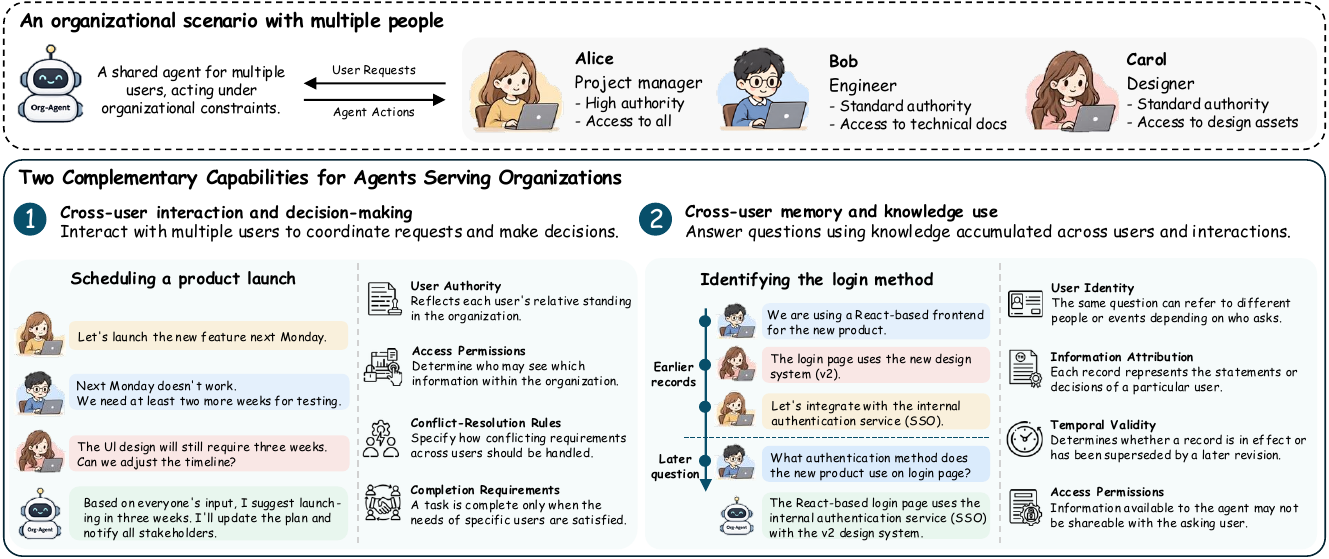}
    \caption{\textbf{Two complementary capabilities of agents for organizations.} Left: \emph{cross-user interaction and decision-making}, illustrated by coordinating a product launch based on multiple users' requests. Right: \emph{cross-user memory and knowledge use}, illustrated by answering a question about the login method using prior interaction history. }
    \label{fig1}
\end{figure}

We identify two complementary capability dimensions for such agents, as illustrated in Figure~\ref{fig1}. \emph{Cross-user interaction and decision-making} involves interacting with multiple users to coordinate their requests and make decisions. For example, scheduling a product launch requires bringing together the manager's proposed timeline and the progress reported by the engineer and designer to develop a joint launch plan. \emph{Cross-user memory and knowledge use} involves organizing and retrieving knowledge accumulated through interactions with different users to answer subsequent questions. For example, answering a question about a product's login method requires referring back to earlier project discussions among team members. Recent benchmarks make these requirements concrete. MUSES-Bench, introduced in \emph{Multi-User Large Language Model Agents}~\citep{yang2026multi}, evaluates agents on tasks involving instruction selection and following, cross-user information access, and meeting scheduling, while GroupMemBench~\citep{yang2026groupmembench} examines memory and knowledge use in multi-user conversations. These studies establish evaluation settings and show that current agents and memory systems still struggle with both capabilities.



Both capabilities are governed by organizational constraints across three aspects. Regarding users, identity concerns who issues a request, authority reflects the user's relative standing in the organization, and access permissions concern who may see which information. Regarding information, attribution identifies whose statements or decisions the information represents, while temporal validity concerns whether a record is still in effect or has been superseded by a later revision. Regarding joint decisions, conflict-resolution rules specify how conflicting requirements across users should be handled, while completion requirements specify which users' needs must be satisfied for the task to be considered complete. Simply concatenating all users' requests and interactions does not explicitly represent or enforce these constraints, while handling each user's requests independently can overlook the dependencies that these constraints induce across users.

To this end, we propose \textsc{Org-Agent}, a constraint-centric reasoning framework with three stages. First, \emph{task decomposition and graph construction} decomposes the task into executable subtasks and organizes them into a directed acyclic graph (DAG)~\citep{thost2021directed}, whose edges encode the dependencies among subtasks, and the graph is updated as needed when new user inputs change the pending subtasks. Second, \emph{dependency-aware scheduling} topologically sorts the graph into an execution order in which every node comes after the nodes it depends on. Third, \emph{constraint-aware execution} invokes tools according to each node's objective and the constraint rules to support decision-making or acquire information from the interaction history. Specifically, we design evidence-acquisition tools that rank records by similarity, filter them by metadata conditions, and traverse explicit links from a given record to retrieve related records, as well as memory-management tools that maintain an execution memory of accumulated information and intermediate results for subsequent nodes. Together, these stages equip \textsc{Org-Agent} with both capabilities within a unified framework, extending a single-user agent to an organizational agent. Our contributions are fourfold:



\begin{itemize}[leftmargin=1.2em]
    \item We identify \emph{cross-user interaction and decision-making}, as well as \emph{cross-user memory and knowledge use} as complementary capabilities of organizational agents, highlighting that both are governed by organizational constraints. This motivates the design of \textsc{Org-Agent}, a unified framework that organizes execution around these constraints and thereby supports both capabilities.
    \item \textsc{Org-Agent} constructs a dynamic task dependency graph, termed TDG, which is a directed acyclic graph whose nodes are atomic subtasks and whose edges encode the dependencies among them, and updates the graph as needed when new user inputs change the pending subtasks.
    \item On top of the constructed graph, \textsc{Org-Agent} schedules the subtasks in dependency order by topological sorting and executes each of them under the constraint rules, with evidence-acquisition tools for retrieving and linking relevant records from the interaction history and memory-management tools for maintaining accumulated execution memory across subtasks.
    \item Experiments on MUSES-Bench and GroupMemBench demonstrate the effectiveness of \textsc{Org-Agent} in cross-user interaction and decision-making as well as cross-user memory and knowledge use, respectively. Ablation studies further show the contributions of dependency-aware scheduling and tool-augmented execution on both benchmarks.
\end{itemize}


\section{Related Work}

Recent work has begun to examine LLM agents in multi-user environments. MUSES-Bench~\citep{yang2026multi} evaluates the ability of a single agent to serve multiple users within a shared interaction setting, while PeopleJoin~\citep{jhamtani2025peoplejoin} studies how agents identify and communicate with relevant users to gather distributed information. In parallel, agent memory systems~\citep{zhang2025survey} have evolved to help agents store and retrieve information from past interactions. Early work such as MemGPT~\citep{packer2023memgpt} manages hierarchical memory to extend the effective context window. Subsequent systems extract and structure salient information from conversations. Mem0~\citep{chhikara2025mem0} consolidates key facts from conversations, while Zep~\citep{rasmussen2025zep} and Hindsight~\citep{latimer2025hindsight} further organize memory into temporal knowledge graphs and structured memory networks, respectively. More recent systems such as LightMem~\citep{fang2026lightmem} and SimpleMem~\citep{liu2026simplemem} improve efficiency through offline consolidation and compact memory representations. Despite this progress, GroupMemBench~\citep{yang2026groupmembench} reveals persistent limitations of existing memory systems in using information from multi-user conversations. Rather than focusing solely on memory storage and retrieval, \textsc{Org-Agent} makes these constraints explicit during execution and supports cross-user interaction and decision-making as well as cross-user memory and knowledge use.

\section{Preliminaries}

We consider an agent for an organization that serves a set of $N$ users $\mathcal{U}=\{u_1,u_2,\ldots,u_N\}$. Interaction proceeds in discrete turns indexed by $t\in\{1,2,\ldots,T\}$. At each turn, users may submit requests, provide information, or respond to the agent. We denote the input from user $u_i$ at turn $t$ by $I_{i,t}$, which is empty if the user provides no input. Let $\mathcal{H}_t$ denote the interaction history before turn $t$, which consists of records of conversations among users and their exchanges with the agent. Given $\mathcal{H}_t$, the current user inputs, and the organizational constraints $\mathcal{R}$, the agent produces a set of actions

\begin{equation}
\mathcal{A}_t=\pi\bigl(\mathcal{H}_t,\{I_{i,t}\}_{u_i\in\mathcal{U}},\mathcal{R}\bigr),
\end{equation}

where $\pi$ denotes the agent policy. Actions may include retrieving information, requesting input from users, and providing responses or task outcomes. The interaction ends once all pending requests are addressed. Next, we describe the interaction process for each of the two capabilities.

\paragraph{Cross-user interaction and decision-making.} The agent handles requests submitted by multiple users
over multiple turns. Here $\mathcal{H}_t$ mainly records the exchanges between the agent and the users in earlier turns of the task. The user inputs and agent responses at turn $t$ are added to the history to form $\mathcal{H}_{t+1}$. Moreover, each input $I_{i,t}$ may contain multiple requests submitted by user $u_i$ at turn $t$.


\paragraph{Cross-user memory and knowledge use.} We formulate this setting as a single-turn question-answering
task, where the only input $I_{i,1}$ is a question
from user $u_i$. Here $\mathcal{H}_1$ mainly consists of earlier conversations among users that precede the question, and $\mathcal{A}_1$ consists of actions that retrieve and examine relevant records from $\mathcal{H}_1$, followed by a single answer to $u_i$.

\section{Method}
\begin{figure}[t]
    \centering
    \includegraphics[width=\textwidth]{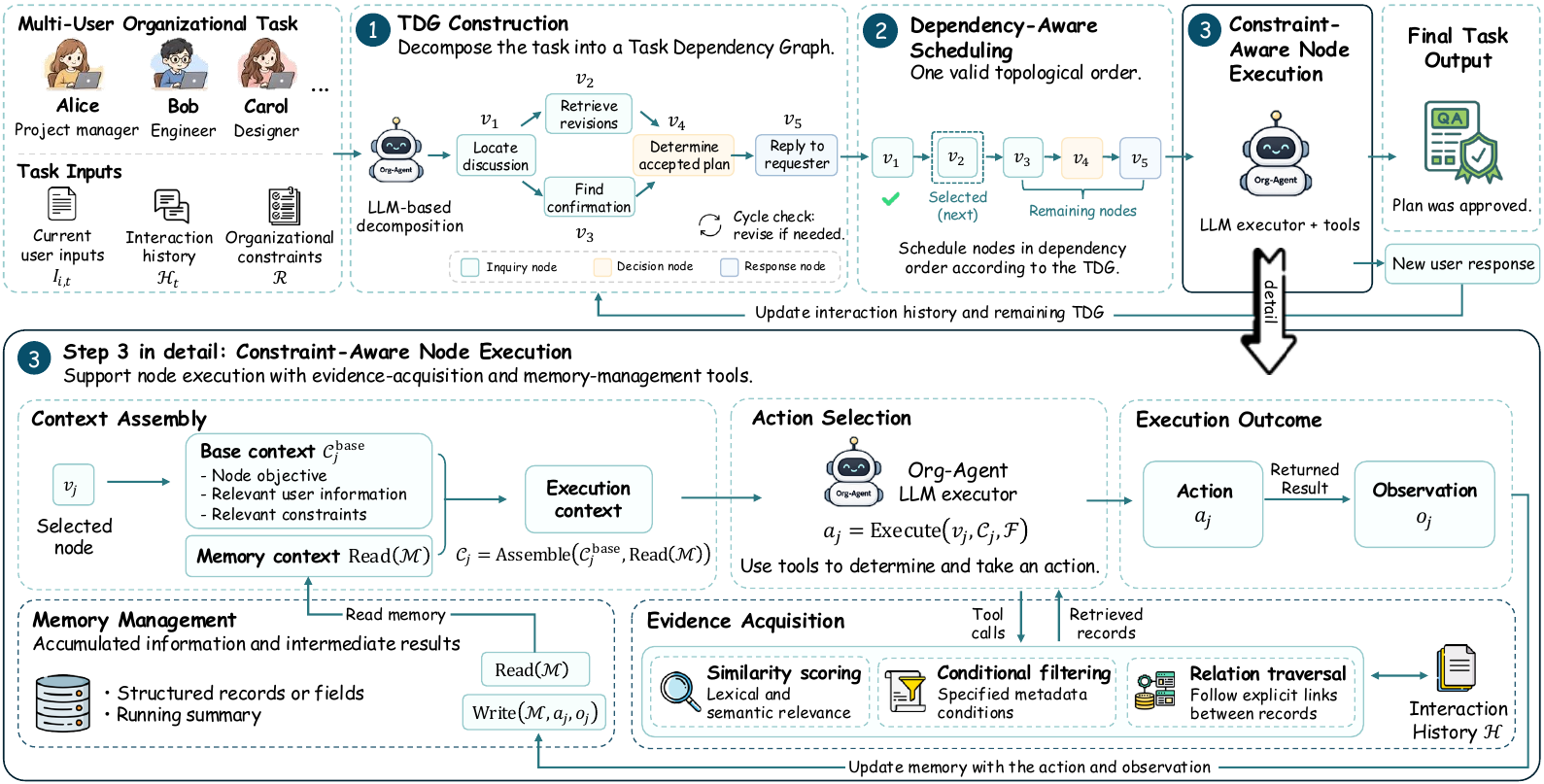}
     \caption{\textbf{Overview of \textsc{Org-Agent}.}
    Given user inputs, interaction history, and organizational constraints,
    \ding{182} \textbf{TDG construction} decomposes the task into subtasks and represents their dependencies in a Task Dependency Graph;
    \ding{183} \textbf{dependency-aware scheduling} orders the nodes through topological sorting;
    and \ding{184} \textbf{constraint-aware node execution} uses the LLM to determine and take actions toward node objectives based on the assembled execution context, with evidence-acquisition tools providing relevant records and memory-management tools supplying memory for context assembly and updating it with information from actions and observations.
   }
    \label{fig:framework}
\end{figure}
\subsection{Overview}

We analyze that the constraints in organizational tasks confront the agent with two fundamental issues. \textbf{(i)~The prerequisites of each action are distributed across users and the interaction history.} For example, a coordination action may require approval from a higher-authority user before it can proceed. Similarly, an information-retrieval action may require identifying the project referred to in a question before retrieving the relevant records. The agent must therefore identify the prerequisites of each action and organize execution according to the resulting dependencies. \textbf{(ii)~Each action must satisfy the relevant constraints during execution.} The agent must therefore determine which constraints in $\mathcal R$ apply to each action and ensure that the action satisfies them. For example, an answer to a question must not disclose information the user is not authorized to access.


Accordingly, our core idea is to decompose a task into executable subtasks and represent their dependencies as a directed acyclic graph. We address (i) by using this graph to execute subtasks in dependency order (Sections~\ref{sec:graph} and~\ref{sec:schedule}), and (ii) by applying the relevant constraint rules in $\mathcal R$ during each subtask's execution, with tools supporting evidence acquisition and memory management (Section~\ref{sec:excute}). The overall framework is illustrated in Figure~\ref{fig:framework}.

\subsection{Task Dependency Graph Construction}
\label{sec:graph}

At turn $t$, we decompose the current pending requests into executable subtasks and organize them into a directed acyclic graph, termed the Task Dependency Graph (TDG). We then describe the details of the graph formulation and construction process below.

\paragraph{Graph Modeling.} Let $\mathcal D_t=\{d_{t,1},\ldots,d_{t,K_t}\}$ denote the set of subtasks at turn $t$, where $K_t$ is the number of subtasks. We represent their dependencies as a directed acyclic graph $\mathcal G_t=(\mathcal V_t,\mathcal E_t)$, where $\mathcal V_t=\{v_{t,1},\ldots,v_{t,K_t}\}$ and each node $v_{t,j}$ corresponds to subtask $d_{t,j}$. Specifically, we distinguish three types of nodes, namely an inquiry node that retrieves relevant records from the interaction history $\mathcal H_t$ or requests information from a user, a decision node that makes a task-specific decision, such as outputting the conclusion of accepting or rejecting a user's instruction, and a response node that replies to a user request. The directed edge set $\mathcal E_t\subseteq\mathcal V_t\times\mathcal V_t$ encodes dependencies among subtasks. An edge $(v_{t,j},v_{t,k})\in\mathcal E_t$ indicates that executing $v_{t,k}$ requires the output of $v_{t,j}$. 





\textbf{Graph Construction.} Building on this formulation, we describe how the framework constructs the graph from the current pending requests in three steps.
\begin{itemize}[leftmargin=1.2em]
    \item \textbf{Subtask decomposition.} An LLM identifies the atomic subtasks needed to fulfill the pending requests from the inputs $\{I_{i,t}\}_{u_i\in\mathcal U}$ and the history $\mathcal H_t$, taking the constraint rules in $\mathcal R$ into account. Each subtask is represented by a concrete objective and associated attributes.
    \item \textbf{Dependency identification.} Given the subtask descriptions and relevant task rules, the LLM identifies which nodes must be resolved before others can be executed and constructs the edge set $\mathcal E_t$ accordingly. Cycles are checked and, if detected, revised by the LLM.
    
    \item \textbf{Graph update.} As nodes are completed, the framework advances through the execution plan. If new user inputs change the pending subtasks, it adjusts the remaining plan, reusing existing dependencies where applicable and reconstructing the graph when needed.
\end{itemize}

\subsection{Dependency-Aware Scheduling}
\label{sec:schedule}

Although the Task Dependency Graph captures dependencies among subtasks, these dependencies must still be translated into an execution order. This stage derives such an order from the acyclic structure of the graph, so that every node is executed only after its prerequisites are available. For brevity, we omit the turn index $t$ from the graph notation hereafter.


\paragraph{Topological Sorting.} Given $\mathcal G=(\mathcal V,\mathcal E)$ with $K$ nodes, we perform a topological sort and index the nodes by their positions in the resulting sequence $\langle v_1,v_2,\ldots,v_K\rangle$, so that
\begin{equation}
(v_j,v_k)\in\mathcal E
\quad\Longrightarrow\quad
j<k.
\end{equation}
This ensures that every prerequisite node precedes the nodes that depend on its output. Nodes can be executed sequentially in this order or organized into topological layers. In layer-wise execution, each layer begins after the preceding layers have completed, and nodes within the same layer may execute in parallel since they have no dependencies on one another.


\subsection{Constraint-Aware Node Execution}
\label{sec:excute}

Rather than relying solely on the model's reasoning to deal with the constraint rules, we support node execution through two categories of predefined tools, namely evidence-acquisition tools and memory-management tools. Let $\mathcal F$ denote the tool set. Given its context $\mathcal C_j$, each node $v_j$ uses the LLM and the available tools in $\mathcal F$ to determine and take an action toward its objective:
\begin{equation}
a_j=\operatorname{Execute}\left(v_j,\mathcal C_j,\mathcal F\right).
\end{equation}
After taking action $a_j$, the node receives the corresponding result as observation $o_j$, and the actions produced by nodes executed at turn $t$ are included in the action set $\mathcal A_t$. The two categories of tools used to support node execution are described below.


\paragraph{Evidence Acquisition.} Evidence-acquisition tools retrieve evidence from $\mathcal H$ and are invoked by the LLM as needed. They include the following three operations.
\begin{itemize}[leftmargin=1.2em]
    \item \textbf{Similarity scoring} takes a textual query and selects candidate records ranked by a combination of lexical and semantic relevance, which identifies useful records related to the node's objective.
    \item \textbf{Conditional filtering} restricts the search to records that satisfy specified metadata conditions, such as author, role, project phase, topic, and channel, so that record selection accounts for contextual conditions as well as relevance to the query.
    \item \textbf{Relation traversal} starts from a specified record and follows explicit links in a chosen direction to retrieve related records, such as replies, the record being replied to, or subsequent decision records, helping trace how a discussion developed and how its decisions changed.
\end{itemize}

During node execution, the LLM selects tools and specifies their arguments based on the node's objective and execution context to refine the search scope or follow links from previously identified records, rather than relying on textual similarity alone.


\paragraph{Memory Management.} The memory-management tool maintains an accumulated execution memory $\mathcal M$ through read and write operations invoked by the framework. It stores task-specific information derived from user interactions together with intermediate results produced during execution. Depending on the task, memory is kept as structured records or fields or as a running summary of the records gathered and conclusions reached so far. Both representations are accessed and updated through the $\operatorname{Read}$ and $\operatorname{Write}$ operations described below.


Before node $v_j$ executes, the read operation provides the available memory for constructing its execution context as
\begin{equation}
\mathcal C_j
=
\operatorname{Assemble}
\left(
\mathcal C_j^{\mathrm{base}},
\operatorname{Read}(\mathcal M)
\right),
\end{equation}
where $\mathcal C_j^{\mathrm{base}}$ contains the node's objective, relevant user information, and relevant constraints from $\mathcal R$. $\operatorname{Read}$ returns the available memory content, including the outputs of the predecessors of $v_j$, and $\operatorname{Assemble}$ combines this information with the base context in the node's prompt template.

After $v_j$ takes its action $a_j\in\mathcal A_t$, the write operation incorporates relevant information from the action and its corresponding observation $o_j$, into memory:
\begin{equation}
\mathcal M\leftarrow\operatorname{Write}(\mathcal M,a_j,o_j),
\end{equation}
where, depending on the memory representation, $\operatorname{Write}$ updates structured records or fields with new information obtained during the current execution, or uses the LLM to revise the running summary. 

\section{Experiments}

To comprehensively evaluate \textsc{Org-Agent}, we design experiments around four questions. \textbf{Q1:} How does \textsc{Org-Agent} compare with baseline methods in the capability of cross-user memory and knowledge use, as evaluated on GroupMemBench? \textbf{Q2:} How does \textsc{Org-Agent} compare with the vanilla baseline in the capability of cross-user interaction and decision-making, as evaluated on MUSES-Bench? \textbf{Q3:} What are the contributions of dependency-aware scheduling and auxiliary tools? \textbf{Q4:} How does performance change as the number of users increases? (Additional experimental evaluations of \textsc{Org-Agent} are presented in Appendix~\ref{app:exp}.)

\subsection{Experimental Setting}

\paragraph{Datasets.} We evaluate \textsc{Org-Agent} on two benchmarks covering the two complementary capabilities. GroupMemBench~\citep{yang2026groupmembench} evaluates cross-user memory and knowledge use through questions grounded in multi-user interaction histories. It contains 745 questions across four domains, namely Finance, Healthcare, Manufacturing, and Technology, and six query categories, namely Multi-Hop, Update, Ambiguity, Implicit, Temporal, and Abstention. MUSES-Bench~\citep{yang2026multi} evaluates cross-user interaction and decision-making through instruction selection (Queue), instruction following (Instruct), cross-user information access (Cross-user Access), and meeting scheduling (Meeting). More details are provided in Appendix~\ref{app:datasets}.

\paragraph{Baselines.}
On GroupMemBench, we compare \textsc{Org-Agent} with two groups of baselines. (i) RAG-based methods include BM25~\citep{robertson2009probabilistic} for sparse lexical retrieval, text-embedding-3-large for dense semantic retrieval, and HippoRAG2~\citep{gutierrez2025rag} and GraphRAG~\citep{edge2024local} for graph-based retrieval. (ii) Memory-based methods include MemGPT~\citep{packer2023memgpt}, LightMem~\citep{fang2026lightmem}, SimpleMem~\citep{liu2026simplemem}, and Hindsight~\citep{latimer2025hindsight}. On MUSES-Bench, we compare \textsc{Org-Agent} with the vanilla agent using the same backbone LLM, treating \textsc{Org-Agent} as a plug-and-play enhancement to the backbone for handling requests from multiple users.

\paragraph{Evaluation Metrics.} On GroupMemBench, we report accuracy for each query category and overall accuracy across all questions, with answer correctness assessed by an LLM judge against reference answers.
On MUSES-Bench, Queue $F_1$ measures the precision--recall balance between the predicted and reference sets of accepted instructions. Instruct accuracy measures the proportion of applicable instruction requirements satisfied by the generated responses, as determined by rule-based checkers. For cross-user access, Privacy measures the proportion of unauthorized users for whom no access violation is detected, while Utility measures the proportion of authorized users judged to have received access to the requested information. For meeting scheduling, success rate measures the proportion of scenarios in which the agent finalizes a schedule that accommodates all required participants within their stated availability windows.

\paragraph{Implementation Details.} On GroupMemBench, all methods use GPT-4o-mini~\citep{hurst2024gpt} for their LLM-based components, while GPT-5.1~\citep{singh2025openai} serves as the judge model. On MUSES-Bench, we evaluate GPT-4o-mini, DeepSeek-V4.1-Flash~\citep{deepseekai2026deepseekv41flashpushinglimitskv}, and Qwen3-32B~\citep{yang2025qwen3}, where each backbone drives both the agent and the simulated users. Multi-turn tasks are limited to at most $T=10$ turns. Further settings are given in Appendix~\ref{app:setting}.

\definecolor{ragheader}{HTML}{D6E9DF}
\definecolor{memoryheader}{HTML}{EAF3EE}

\begin{table*}[t]
    \centering
    \setlength{\belowcaptionskip}{8pt}
    \caption{\textbf{Results (\%) of baselines and \textsc{Org-Agent} on GroupMemBench across six query categories using GPT-4o-mini.} The best result for each category is highlighted in \textbf{bold}, while the second best is indicated with an \underline{underline}. Avg. denotes the average performance.}
    \label{tab:groupmembench}
    \renewcommand{\arraystretch}{1.3}
    \resizebox{\textwidth}{!}{%
    \begin{tabular}{lccccccc}
        \toprule
        \textbf{Method}
        & \textbf{Multi-Hop}
        & \textbf{Update}
        & \textbf{Ambiguity}
        & \textbf{Implicit}
        & \textbf{Temporal}
        & \textbf{Abstention}
        & \textbf{Avg.} \\
        \midrule
        \rowcolor{ragheader}
        \multicolumn{8}{l}{\textit{RAG-based Methods}} \\
        BM25
        & 36.26 & 15.89 & 14.15 & \underline{42.86} & 41.36 & 68.35 & 37.72 \\
        text-embedding-3-large
        & 32.42 & 12.15 & 23.58 & 34.69 & 25.31 & 66.91 & 33.29 \\
        HippoRAG2
        & 28.57 & \underline{18.69} & 14.15 & 34.69 & 33.33 & \textbf{79.86} & 36.11 \\
        GraphRAG
        & 8.24 & 0.93 & 6.60 & 14.29 & 1.85 & \underline{78.42} & 19.06 \\
        \midrule
        \rowcolor{memoryheader}
        \multicolumn{8}{l}{\textit{Agent Memory Systems}} \\
        MemGPT
        & 24.18 & 10.28 & 23.58 & 30.61 & 14.20 & 60.43 & 27.11 \\
        Hindsight
        & \underline{43.41} & 10.28 & \textbf{33.96} & 38.78 & \textbf{44.44} & 51.80 & \underline{38.79} \\
        LightMem
        & 6.04 & 0.00 & 5.66 & 6.12 & 3.70 & 72.66 & 17.05 \\
        SimpleMem
        & 23.63 & 7.48 & 16.98 & 20.41 & 24.69 & 69.78 & 28.99 \\
        \midrule
        \textsc{Org-Agent} (Ours)
        & \textbf{49.45} & \textbf{28.97} & \underline{25.47} & \textbf{48.98} & \underline{42.59} & 62.59 & \textbf{44.03} \\
        \bottomrule
    \end{tabular}%
    }
\end{table*}


\subsection{Performance on Cross-User Memory and Knowledge Use (\Q)}

To address Q1, we conduct a comprehensive comparison of various baseline methods with \textsc{Org-Agent} on GroupMemBench. The detailed experimental results are presented in Table \ref{tab:groupmembench}. Based on our analysis, we derive the following key observations.

\obs{Existing agent memory systems do not consistently outperform basic retrieval baselines on cross-user memory and knowledge use.} Although these systems are designed to organize and reuse interaction history, most of them fall behind BM25 in average accuracy. Specifically, SimpleMem and MemGPT reach 28.99\% and 27.11\%, falling behind BM25 by 8.73 and 10.61 points, and LightMem drops to 17.05\%. We analysis that compact memory representations omit contextual details needed to interpret and select relevant records in cross-user tasks. Moreover, although Hindsight relies on costly memory construction, it exceeds BM25 by only 1.07 points.

\obs{\textsc{Org-Agent} demonstrates improved cross-user memory and knowledge utilization capabilities compared with both retrieval baselines and agent memory systems.} On GroupMemBench, \textsc{Org-Agent} attains the best average accuracy among all compared methods. Specifically, it reaches 44.03\%, exceeding BM25 and text-embedding-3-large by 6.31 and 10.74 points, and the strongest memory system, Hindsight, by 5.24 points. Taken together, these comparisons support the effectiveness of the approach across the two baseline groups. Notably, its gains do not require constructing a memory graph before the query is received.



\subsection{Performance on Cross-User Interaction and Decision-Making (\Q)}

To analyze the performance in cross-user interaction and decision-making scenarios, we compare \textsc{Org-Agent} with the vanilla baseline on MUSES-Bench across four tasks and three backbones. The results in Table~\ref{tab:muses_results} yield the following observations.

\obs{\textsc{Org-Agent} improves cross-user interaction and decision-making capability over the vanilla baseline.} On MUSES-Bench, \textsc{Org-Agent} raises the average score with every backbone we evaluate. Specifically, with GPT-4o-mini it reaches 72.39\% on average, exceeding the vanilla baseline by 8.27 points, with Instruct accuracy rising from 58.69\% to 73.41\% and Meeting success rate from 37.04\% to 60.19\%. The largest gains appear on Meeting, reaching 23.15 percentage points with GPT-4o-mini, as this task most clearly illustrates how constraints of organizational tasks affect task completion through interactions among users over multiple turns.

\obs{The improvements hold across backbones of different capabilities.} \textsc{Org-Agent} improves the average score with all three backbones by 8.27 points with GPT-4o-mini, 4.24 points with DeepSeek-V4.1-Flash, and 5.29 points with Qwen3-32B. The gain is larger for the two weaker backbones, whose vanilla scores are around 64, than for DeepSeek-V4.1-Flash, whose vanilla score is 80.8, which suggests that explicit constraints handling compensates for weaker reasoning in the backbone. Moreover, these backbones span proprietary and open-source models at different scales, supporting the generality of \textsc{Org-Agent} as a model-agnostic framework.


\definecolor{oursrow}{HTML}{EAF3EE}

\begin{table*}[t]
\centering
\setlength{\belowcaptionskip}{8pt}
\caption{\textbf{Results (\%) of Vanilla and \textsc{Org-Agent} on MUSES-Bench across three backbones.}
The best result for each metric within each backbone is shown in \textbf{bold}.
Avg. denotes the average
performance. $\Delta \uparrow$ denotes the difference between \textsc{Org-Agent} and Vanilla in percentage points.}
\label{tab:muses_results}
\renewcommand{\arraystretch}{1.3}
\resizebox{\textwidth}{!}{%
\begin{tabular}{llcccccc}
\toprule
\multirow{2}{*}{\textbf{Model}}
& \multirow{2}{*}{\textbf{Method}}
& \multicolumn{2}{c}{\textbf{Multi-user Instruction}}
& \multicolumn{2}{c}{\textbf{Cross-user Access}}
& \textbf{Meeting}
& \multirow{2}{*}{\textbf{Avg.}} \\
\cmidrule(lr){3-4}
\cmidrule(lr){5-6}
\cmidrule(lr){7-7}
& & Queue ($F_1$) & Instruct (Acc.)
& Privacy & Utility & Success Rate & \\
\midrule
& Vanilla
& 62.97 & 58.69 & 96.94 & \textbf{64.96} & 37.04 & 64.12 \\
& \textsc{Org-Agent} (Ours)
& \textbf{74.24} & \textbf{73.41} & \textbf{97.59}
& 56.52 & \textbf{60.19} & \textbf{72.39} \\
\rowcolor{oursrow}
\cellcolor{white}\multirow{-3}{*}{GPT-4o-mini}
& $\Delta \uparrow$
& $+11.27$ & $+14.72$ & $+0.65$
& $-8.44$ & $+23.15$ & $+8.27$ \\
\midrule
& Vanilla
& 78.16 & 81.18 & 71.26 & 97.35 & 75.93 & 80.78 \\
& \textsc{Org-Agent} (Ours)
& \textbf{88.21} & \textbf{83.78} & \textbf{75.25}
& \textbf{99.17} & \textbf{78.70} & \textbf{85.02} \\
\rowcolor{oursrow}
\cellcolor{white}\multirow{-3}{*}{DeepSeek-V4.1-Flash}
& $\Delta \uparrow$
& $+10.05$ & $+2.60$ & $+3.99$
& $+1.82$ & $+2.77$ & $+4.24$ \\
\midrule
& Vanilla
& 62.38 & 63.46 & \textbf{64.28} & 91.10 & 36.11 & 63.47 \\
& \textsc{Org-Agent} (Ours)
& \textbf{64.67} & \textbf{81.77} & 47.45
& \textbf{98.05} & \textbf{51.85} & \textbf{68.76} \\
\rowcolor{oursrow}
\cellcolor{white}\multirow{-3}{*}{Qwen3-32B}
& $\Delta \uparrow$
& $+2.29$ & $+18.31$ & $-16.83$
& $+6.95$ & $+15.74$ & $+5.29$ \\
\bottomrule
\end{tabular}%
}
\end{table*}

\subsection{Ablation Studies (\Q)}

\begin{wrapfigure}{t}{0.5\textwidth}
    \centering
    \includegraphics[width=0.48\textwidth]{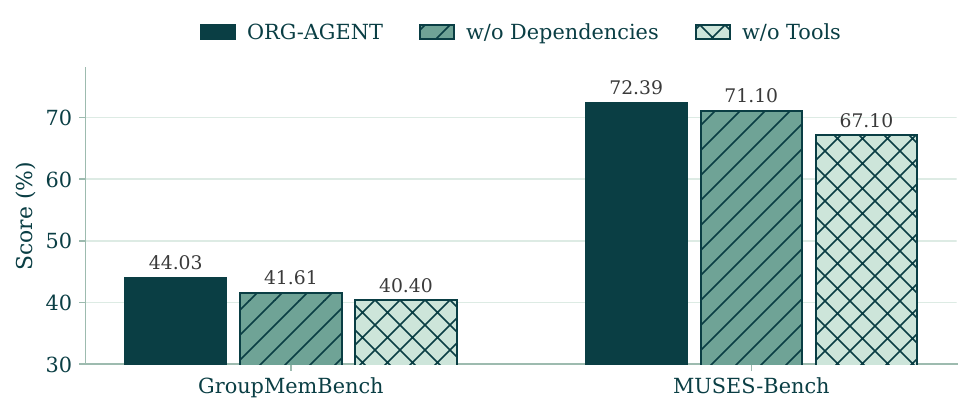}
    \caption{\textbf{Ablation study on key components of \textsc{Org-Agent} across GroupMemBench and MUSES-Bench with GPT-4o-mini.} The $y$-axis represents the average score of the full model and its two variants on each benchmark.}
    \label{fig:ablation}
\end{wrapfigure}

To address Q3, we conduct ablation studies on the two core components of \textsc{Org-Agent} across GroupMemBench and MUSES-Bench, using GPT-4o-mini as the backbone in all cases. We consider two variants: (i) \textit{w/o Dependencies} removes all edges of the TDG, so that every subtask is executed independently without topological order. (ii) \textit{w/o Tools} keeps the TDG and its execution order but disables conditional filtering, relation traversal, and memory management, so that each node retrieves records by similarity scoring alone and receives the unprocessed results of earlier nodes instead of the organized execution memory. Results on both benchmarks are reported in Figure~\ref{fig:ablation}.

\obs{Both dependency-aware scheduling and constraint-aware node execution contribute to the effectiveness of \textsc{Org-Agent}.} Removing either component reduces the average score on both benchmarks. On GroupMemBench, the average score drops from 44.03\% to 41.61\% without dependencies and to 40.40\% without tools. On MUSES-Bench, it decreases from 72.39\% to 71.10\% and 67.10\%, respectively. These results highlight the contributions of dependency-aware scheduling and tool-supported node execution to performance on multi-user tasks.

\subsection{Performance across Different User Group Sizes (\Q)}

To address this question, we examine how the size of the user group affects performance on MUSES-Bench. We report evaluation results for three representative tasks, namely Queue, Instruct, and Meeting. Using GPT-4o-mini as the backbone for both \textsc{Org-Agent} and the vanilla baseline, we then compare their scores as the number of users grows. 

\begin{figure}[t]
    \centering
    \includegraphics[width=\textwidth]{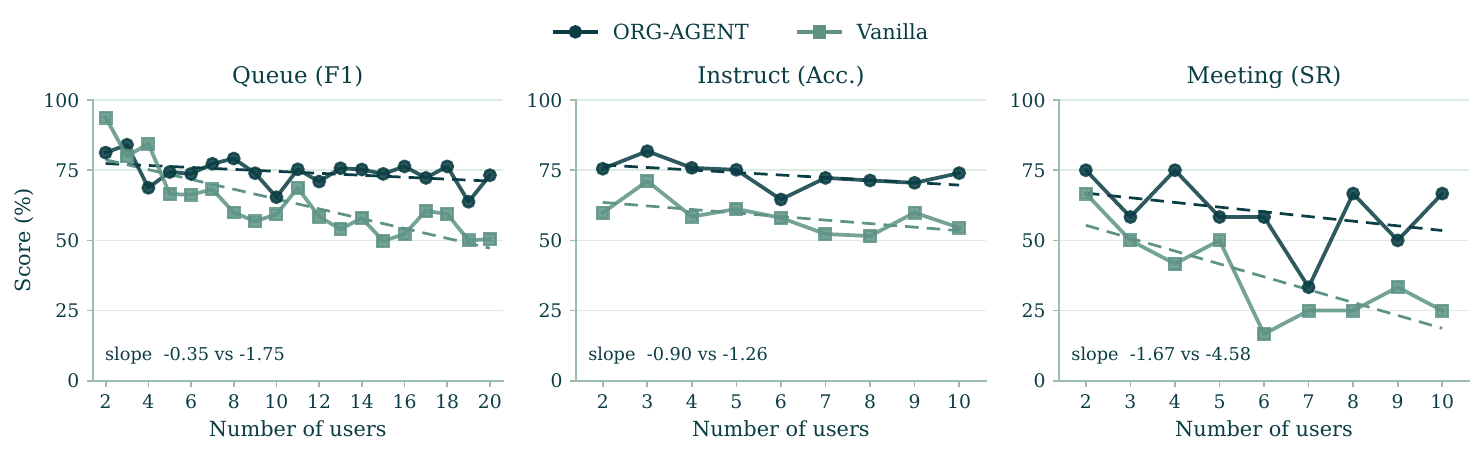}
    \caption{\textbf{Performance across different user group sizes on MUSES-Bench using GPT-4o-mini.} The $x$-axis represents the number of users, and the $y$-axis represents the metric of each task. Dashed lines show a least-squares trend weighted by the number of scenarios at each group size.}
    \label{fig:groupsize}
\end{figure}

\obs{\textsc{Org-Agent} degrades more slowly than the vanilla baseline as the number of users grows.} Both methods decline as groups become larger, but at different rates, shown by the dashed trends in Figure~\ref{fig:groupsize}. The dashed lines are fitted by least squares, weighted by the number of scenarios at each group size. Specifically, \textsc{Org-Agent} loses 0.35 points per additional user on Queue and 1.67 on Meeting, whereas the vanilla baseline loses 1.75 and 4.58. Instruct follows the same pattern with a smaller difference, 0.90 against 1.26. The fitted trends therefore indicate a widening performance gap as the number of users increases within the evaluated range. As the user group grows, the agent must account for requests and constraints involving more users. The slower performance decline suggests that \textsc{Org-Agent} is better able to handle this increased coordination burden within the evaluated range of group sizes.


\section{Conclusion}

We presented \textsc{Org-Agent}, a unified framework for agents serving multiple users in organizational environments. It addresses two complementary capabilities, namely cross-user interaction and decision-making, as well as cross-user memory and knowledge use. Both are governed by organizational constraints on users, information, and joint decisions that shape the prerequisites of an action and the conditions that must be satisfied during its execution. \textsc{Org-Agent} makes the former explicit in a Task Dependency Graph that is updated as needed when new user inputs arrive, schedules the subtasks in dependency order by topological sorting, and accounts for the latter during each node's execution, with support from evidence-acquisition and memory-management tools. Experiments on GroupMemBench and MUSES-Bench demonstrate the effectiveness of this approach on both capabilities. On GroupMemBench, \textsc{Org-Agent} improves overall accuracy over the baselines. On MUSES-Bench, it improves average performance across all three evaluated backbones. Ablation studies support the contributions of dependency modeling and tool-supported execution, while analyses of different user group sizes show slower performance degradation than the vanilla agent as the number of users grows in the evaluated settings. These findings highlight the value of organizing task execution around organizational constraints, and suggest a path for LLM agents to move beyond personal assistants toward organizational agents.

\subsection*{Reproducibility Statement}
All code and relevant resources can be accessed from the anonymous repository referenced in the abstract. A thorough README document within the repository supplies full guidance to reproduce our experimental setups and outcomes.

\subsection*{AI Use Statement}
\label{app:ai_use}

We used AI tools to assist with grammar checking and language refinement. The authors take full responsibility for the accuracy and integrity of the final manuscript.

\bibliography{iclr2027_conference}
\bibliographystyle{iclr2027_conference}

\appendix
\newpage
\section{Datasets}
\label{app:datasets}

We conduct our experiments on two benchmarks corresponding to the two capabilities studied in this paper. GroupMemBench evaluates cross-user memory and knowledge use through question answering over multi-user conversations, while MUSES-Bench evaluates cross-user interaction and decision-making through tasks involving a shared agent and multiple users. Our evaluation includes 745 questions from GroupMemBench and 1{,}183 scenarios from MUSES-Bench. For MUSES-Bench, we evaluate four tasks and use the partial-disclosure subset for Meeting, with group sizes ranging from 2 to 20 users across tasks. Table~\ref{tab:dataset_statistics} summarizes the statistics of both benchmarks.

\begin{wraptable}{r}{0.5\textwidth}
    \centering
    \setlength{\belowcaptionskip}{8pt}
    \small
    \caption{\textbf{Statistics of the benchmarks in evaluation.} Samples correspond to questions for GroupMemBench and scenarios for MUSES-Bench.}
    \label{tab:dataset_statistics}
    \begin{tabular}{llr}
        \toprule
        \textbf{Benchmark} & \textbf{Category} & \textbf{\# Samples} \\
        \midrule
        GroupMemBench
        & Multi-Hop         & 182 \\
        & Update            & 107 \\
        & Ambiguity         & 106 \\
        & Implicit          & 49 \\
        & Temporal          & 162 \\
        & Abstention        & 139 \\
        \cmidrule(lr){2-3}
        & \textbf{Total}    & \textbf{745} \\
        \midrule
        MUSES-Bench
        & Queue             & 304 \\
        & Instruct          & 555 \\
        & Cross-user Access & 216 \\
        & Meeting           & 108 \\
        \cmidrule(lr){2-3}
        & \textbf{Total}    & \textbf{1,183} \\
        \bottomrule
    \end{tabular}
\end{wraptable}

\textbf{GroupMemBench} provides conversation histories annotated with user identities and timestamps, with each question associated with an asking user. The questions cover six categories. \textit{Multi-Hop} requires linking evidence across multiple messages or threads. \textit{Update} tests the tracking of revised facts and decisions, including their current values, previous values, or reasons for change. \textit{Ambiguity} requires interpreting role-dependent terminology and connecting expressions used by different speakers. \textit{Implicit} requires resolving first-person references using the asking user's identity and history. \textit{Temporal} tests reasoning about event timing and ordering using timestamps and relative time expressions. \textit{Abstention} requires recognizing that the requested information is absent from the history rather than fabricating an answer. These categories contain 182, 107, 106, 49, 162, and 139 questions, respectively. 

\textbf{MUSES-Bench} evaluates a shared agent under task-specific objectives, authority rules, and information-access constraints. \textit{Queue} contains 304 scenarios with 2 to 20 users. The agent must select which instructions to accept according to their alignment with the global objective and the users' authority hierarchy, rejecting instructions that violate the objective or lose a conflict under these rules. \textit{Instruct} contains 555 scenarios with 2 to 10 users, comprising 187 aligned and 368 conflict scenarios. User requests include automatically verifiable requirements, such as length and wording constraints.
In aligned scenarios, the agent should satisfy all users' requirements; in conflict scenarios, evaluation focuses on those issued by the highest-authority users. \textit{Cross-user Access} contains 216 scenarios with 2 to 10 users. Each scenario specifies a protected resource and an authorized-user set. The agent must support authorized access while preventing unauthorized disclosure, with performance evaluated through Privacy and Utility scores. \textit{Meeting} uses 108 partial-disclosure scenarios with 2 to 10 users. Participants have preferred and secondary availability, which may need to be elicited through interaction. The agent must gather the necessary information, negotiate scheduling conflicts, and finalize a time slot that all required participants can attend.

\section{Implementation Details}
\label{app:setting}

\begin{table}[h]
    \centering
    \setlength{\belowcaptionskip}{8pt}
    \caption{Hardware configuration of our local experimental server.}
    \begin{tabular}{ll}
        \toprule
        \textbf{Component} & \textbf{Specification} \\
        \midrule
        GPU & $4 \times$ NVIDIA GeForce RTX 4090 (24 GB each) \\
            & $4 \times$ NVIDIA GeForce RTX 4090 D (24 GB each) \\
        CPU & $2 \times$ Intel Xeon Gold 6426Y (16 cores each) \\
        \bottomrule
    \end{tabular}

    \label{tab:hardware}
\end{table}

\paragraph{Hardware.}
The hardware configuration of our local experimental server is
summarized in Table~\ref{tab:hardware}.

\paragraph{Prompting settings.} On GroupMemBench, all baselines use the same question-answering prompt template to ensure a consistent evaluation setup. On MUSES-Bench, all methods share the same task-specific system instructions and user-simulation prompts.

\section{Details of Baselines}
\label{app:baselines}

We provide additional descriptions of the baseline methods used in our experiments. On GroupMemBench, the baselines cover RAG-based methods and agent memory systems, representing different approaches to accessing and organizing information from interaction histories. On MUSES-Bench, we compare against a vanilla agent with the same backbone LLM to examine the contribution of the proposed framework beyond the backbone's existing capabilities.

\subsection{RAG-Based Methods}

Retrieval-based methods identify information relevant to a question and provide it as context for answer generation~\citep{lewis2020retrieval}. The selected baselines cover sparse lexical retrieval, dense semantic retrieval, and retrieval over graph-structured representations.

\paragraph{BM25.}
BM25~\citep{robertson2009probabilistic} is a sparse retrieval method that ranks records according to their lexical relevance to a query. Its scoring function combines term frequency, inverse document frequency, and document-length normalization. 


\paragraph{text-embedding-3-large.}
This baseline uses text-embedding-3-large to encode queries and historical records into dense vector representations. Records are ranked by similarity in the embedding space and supplied as evidence for answering the question. 


\paragraph{HippoRAG 2.}
HippoRAG~2~\citep{gutierrez2025rag} combines graph-based retrieval with dense representations to support access to interconnected information. It builds on Personalized PageRank and incorporates passage information into the retrieval structure, while using an LLM during query processing to improve evidence selection. This baseline evaluates whether graph-based associations can help recover information distributed across multiple records, beyond what can be obtained through direct lexical or semantic matching alone.

\paragraph{GraphRAG.}
GraphRAG~\citep{edge2024local} uses an LLM to extract entities and relationships from source documents and organize them into a graph index. It also constructs summaries of graph communities to support answering questions over related information.


\subsection{Agent Memory Systems}

Agent memory systems maintain representations derived from previous interactions and make this information available for subsequent queries. The selected methods differ in how they manage memory, consolidate information, and retrieve relevant content. 

\paragraph{MemGPT.}
MemGPT~\citep{packer2023memgpt} introduces a hierarchical memory-management architecture inspired by operating systems. It manages information across memory tiers and transfers relevant content into the LLM's limited context window when needed. This design supports interactions whose accumulated history exceeds the available context. 


\paragraph{LightMem.}
LightMem~\citep{fang2026lightmem} is a memory framework that separates online information processing from offline consolidation. It first filters and groups incoming information, then organizes related content into compact memory representations. An offline update process further consolidates long-term memory without placing all maintenance operations on the inference path. 

\paragraph{SimpleMem.}
SimpleMem~\citep{liu2026simplemem} constructs compact memory units through semantic structured compression and integrates related information through online semantic synthesis, merging related context within a session to reduce redundancy. At query time, its intent-aware retrieval planning determines what information to retrieve and how broadly to search.

\paragraph{Hindsight.}
Hindsight~\citep{latimer2025hindsight} organizes memory into distinct logical networks for world facts, agent experiences, entity summaries, and evolving beliefs. Its retain, recall, and reflect operations support information ingestion, retrieval, and reasoning over the resulting memory bank. The architecture incorporates temporal and entity information to support the interpretation of accumulated knowledge. It provides a structured-memory baseline for evaluating questions that require connecting and distinguishing information across users and interactions.

\subsection{Vanilla Agent on MUSES-Bench}

The vanilla baseline uses the backbone LLM within the benchmark's original interaction protocol, without the Task Dependency Graph or the execution tools introduced by \textsc{Org-Agent}. It receives the task instructions and user inputs and generates responses or decisions directly. For multi-turn tasks, it can continue interacting with users through the benchmark environment rather than being restricted to a single response. We compare the vanilla agent and \textsc{Org-Agent} using the same backbone LLM under each of the three evaluated backbones.


\section{Task-Specific Node Examples}
\label{app:node_examples}

The content of a node depends on the task being performed. On GroupMemBench, a node may retrieve records relevant to a user's question. On MUSES-Bench, nodes may evaluate an instruction, generate a response, handle an information-access request, or ask a participant for confirmation. Table~\ref{tab:node_examples} illustrates these task-specific node objectives. The examples describe individual subtasks rather than complete task workflows.

\begin{table}[htbp]
    \centering
    \setlength{\belowcaptionskip}{8pt}
    \caption{\textbf{Examples of task-specific nodes in \textsc{Org-Agent} across GroupMemBench and MUSES-Bench.} Each node represents a subtask with a concrete objective.}
    \label{tab:node_examples}
    \begin{tabularx}{\linewidth}{lX}
        \toprule
        \textbf{Benchmark / Task} & \textbf{Example node} \\
        \midrule
        GroupMemBench
        & Find User\_13's decision about the UAT validation scope \\
        \midrule
        MUSES-Bench: Queue
        & Decide whether to accept Alice's instruction to prioritize urgent requests \\
        \midrule
        MUSES-Bench: Instruct
        & Respond to Alice in English using at most three sentences \\
        \midrule
        MUSES-Bench: Cross-user Access
        & Handle Bob's request for the project budget \\
        \midrule
        MUSES-Bench: Meeting
        & Ask David whether he can attend on Friday at 14:00 \\
        \bottomrule
    \end{tabularx}
\end{table}

\section{Additional Experiments}
\label{app:exp}

To further analyze \textsc{Org-Agent}, we conduct additional experiments and case studies addressing four questions. \textbf{Q5:} Does \textsc{Org-Agent} remain effective on GroupMemBench with a different backbone model? \textbf{Q6:} How sensitive is its performance to the hybrid-retrieval weight $\alpha$ that balances lexical and semantic relevance? \textbf{Q7:} How much LLM cost does \textsc{Org-Agent} incur compared with the baselines, and how does this cost relate to accuracy? \textbf{Q8:} How do constraint-aware node execution and dependency-aware scheduling operate in concrete multi-user cases?

\subsection{Effectiveness of different backbones (\Q)}

In this section, we study the effectiveness of \textsc{Org-Agent} across different model backbones. Since the main experiments on GroupMemBench use GPT-4o-mini, we additionally evaluate \textsc{Org-Agent} and the baselines with Qwen3-8B as the backbone model to assess whether its performance advantage over these baseline methods extends to another backbone.

\obs{\textsc{Org-Agent} remains effective with a different model backbone.}
As shown in Table~\ref{tab:groupmembench_qwen3_8b}, \textsc{Org-Agent} achieves an overall accuracy of 40.40\% with Qwen3-8B, outperforming the strongest baseline, BM25, by 6.04 percentage points and the strongest memory system, Hindsight, by 7.65 percentage points. These results show that its overall advantage is maintained with Qwen3-8B, providing evidence that the effectiveness of the framework is not limited to GPT-4o-mini.


\begin{table*}[t]
    \centering
    \setlength{\belowcaptionskip}{8pt}
    \caption{\textbf{Results (\%) of baselines and \textsc{Org-Agent} on GroupMemBench across six query categories using Qwen3-8B.} The best result for each category is highlighted in \textbf{bold}, while the second best is indicated with an \underline{underline}. Avg. denotes the average performance.}
    \label{tab:groupmembench_qwen3_8b}
    \renewcommand{\arraystretch}{1.3}
    \resizebox{\textwidth}{!}{%
    \begin{tabular}{lccccccc}
        \toprule
        \textbf{Method}
        & \textbf{Multi-Hop}
        & \textbf{Update}
        & \textbf{Ambiguity}
        & \textbf{Implicit}
        & \textbf{Temporal}
        & \textbf{Abstention}
        & \textbf{Avg.} \\
        \midrule
        \rowcolor{ragheader}
        \multicolumn{8}{l}{\textit{RAG-based Methods}} \\
        BM25
        & \underline{36.81} & \textbf{28.04} & 18.87 & 36.73 & \underline{38.89} & 41.73 & \underline{34.36} \\
        text-embedding-3-large
        & 29.12 & 22.43 & 22.64 & \underline{42.86} & 17.28 & 33.09 & 26.31 \\
        HippoRAG2
        & 29.67 & 23.36 & 17.92 & 38.78 & 24.07 & \textbf{64.03} & 32.89 \\
        GraphRAG
        & 7.69 & 0.93 & 5.66 & 6.12 & 4.32 & 51.80 & 13.83 \\
        \midrule
        \rowcolor{memoryheader}
        \multicolumn{8}{l}{\textit{Agent Memory Systems}} \\
        MemGPT
        & 20.33 & 14.95 & 20.75 & 32.65 & 10.49 & 35.97 & 21.21 \\
        Hindsight
        & 34.07 & 11.21 & \underline{28.30} & \textbf{48.98} & 38.27 & 38.85 & 32.75 \\
        LightMem
        & 16.48 & 3.74 & 15.09 & 14.29 & 13.58 & 40.29 & 18.12 \\
        SimpleMem
        & 25.27 & 14.02 & 13.21 & 14.29 & 29.01 & 46.04 & 25.91 \\
        \midrule
        \textsc{Org-Agent} (Ours)
        & \textbf{39.01} & \underline{25.23} & \textbf{34.91} & 34.69 & \textbf{39.51} & \underline{61.15} & \textbf{40.40} \\
        \bottomrule
    \end{tabular}%
    }
\end{table*}

\subsection{Hyper-parameter Sensitivity (\Q)}

\begin{wrapfigure}{r}{0.5\textwidth}
    \centering
    \includegraphics[width=0.48\textwidth]{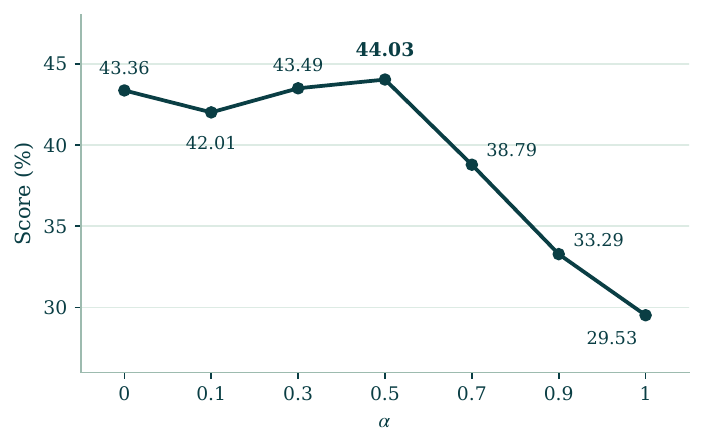}
    \caption{\textbf{Effect of the hybrid-retrieval weight $\alpha$ on GroupMemBench with GPT-4o-mini.} The $x$-axis is $\alpha$, and the $y$-axis is the weighted average accuracy.}
    \label{fig:alpha_sensitivity}
\end{wrapfigure}

In this section, we examine the sensitivity of \textsc{Org-Agent} to the weight $\alpha$ in the similarity scoring tool introduced in Section~\ref{sec:excute}, evaluated on GroupMemBench using GPT-4o-mini. This tool ranks candidate records by combining BM25 lexical relevance and dense semantic similarity computed with text-embedding-3-large,  and $\alpha$ controls their relative contribution. For a retrieval query $x$ and a candidate record $h$ in the interaction history $\mathcal H$, the combined score is defined as
\begin{equation}
\operatorname{Score}(x,h)
=
(1-\alpha)\,\widetilde{s}_{\mathrm{BM25}}(x,h)
+
\alpha\,\widetilde{s}_{\mathrm{dense}}(x,h),
\end{equation}
where $\widetilde{s}_{\mathrm{BM25}}$ and $\widetilde{s}_{\mathrm{dense}}$ denote the BM25 score and the dense cosine similarity, respectively, each min-max normalized over the current candidate set. Thus, $\alpha=0$ ranks records by lexical relevance only, while $\alpha=1$ ranks them by semantic similarity only. Figure~\ref{fig:alpha_sensitivity} reports the average score for $\alpha\in\{0,0.1,0.3,0.5,0.7,0.9,1.0\}$.

\obs{Balanced lexical and semantic weighting achieves the best performance among the evaluated settings.} \textsc{Org-Agent} achieves its highest average score at $\alpha=0.5$, while larger dense weights lead to lower performance. Specifically, the average score reaches 44.03 at $\alpha=0.5$, compared with 43.36 at $\alpha=0$ and 29.53 at $\alpha=1$. These results suggest that lexical relevance remains important for evidence selection in this setting, supporting our default choice of $\alpha=0.5$.

\subsection{Cost Analysis (\Q)}

\begin{wrapfigure}{r}{0.5\textwidth}
    \centering
    \includegraphics[width=\linewidth]{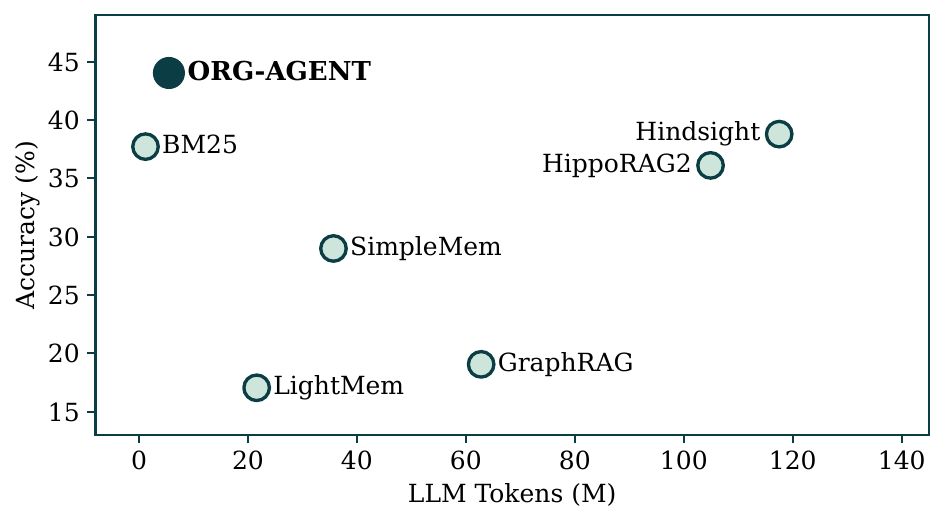}
    \caption{\textbf{LLM cost and accuracy on GroupMemBench with GPT-4o-mini.} The $x$-axis is the number of LLM tokens consumed in the build and question-answering stages, and the $y$-axis is the average accuracy.}
    \label{fig:cost}
\end{wrapfigure}

To address Q7, we measure the LLM tokens consumed by \textsc{Org-Agent} and the baselines on GroupMemBench using GPT-4o-mini, covering both the build stage and the question-answering stage. Figure~\ref{fig:cost} plots the number of tokens over against accuracy.

\obs{\textsc{Org-Agent} delivers the highest accuracy at moderate LLM cost.} As shown in Figure~\ref{fig:cost}, \textsc{Org-Agent} achieves the highest accuracy of 44.03\% while consuming only 5.5M tokens, which is 3.9 times fewer than LightMem (21.6M) and 21.4 times fewer than the strongest baseline, Hindsight (117.5M). Although BM25 is cheaper at 1.2M tokens, it reaches only 37.72\% because it selects records by lexical matching alone. Among the remaining baselines, higher accuracy generally comes at a higher cost, as HippoRAG2 and Hindsight consume more than 100M tokens to reach 36.11\% and 38.79\%, respectively. Yet more tokens do not guarantee better accuracy, since GraphRAG consumes 62.8M tokens but reaches only 19.06\%, below SimpleMem with 35.7M tokens. \textsc{Org-Agent} departs from this trend because it does not construct memories or graphs over the entire history in advance, and instead resolves each question through a Task Dependency Graph at query time.


\subsection{Case Study (\Q)}

\begin{figure}[t]
    \centering
    \includegraphics[width=\textwidth]{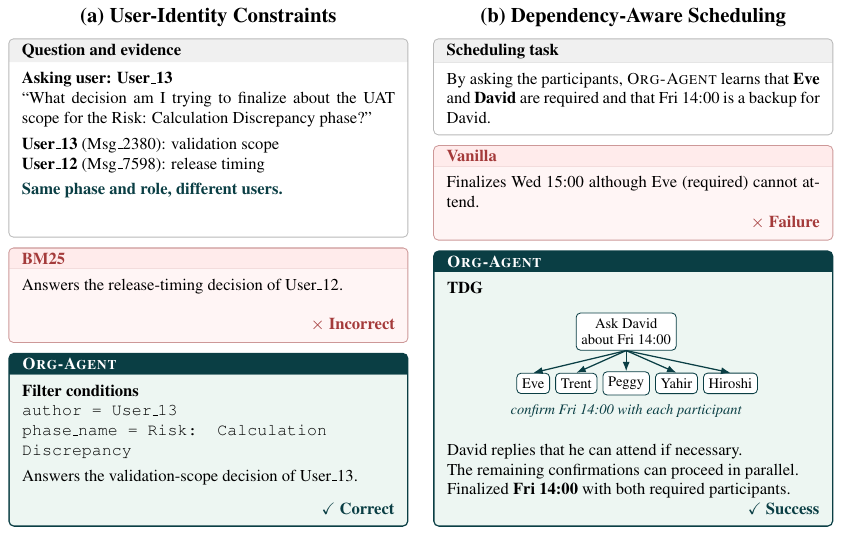}
    \caption{\textbf{Case studies on GroupMemBench and MUSES-Bench.} (a) \textsc{Org-Agent} identifies the asking user's decision rather than another user's related decision by filtering records according to the user's identity and the relevant project phase. (b) With David identified as a required participant, the \textsc{Org-Agent} queries him before proposing the slot to the others, and his confirmation allows the remaining confirmation requests to proceed.}
    \label{fig:case_study}
\end{figure}

To illustrate how \textsc{Org-Agent} behaves on multi-user tasks, we examine two cases in Figure~\ref{fig:case_study}. The first case, from GroupMemBench, shows how constraint-aware node execution uses the asking user's identity to guide record selection through tools. The second case, from MUSES-Bench, shows how dependency-aware scheduling orders interactions with different users.

\obs{Constraint-aware node execution uses the asking user's identity to select relevant records.} In Figure~\ref{fig:case_study}~(a), User\_13 asks which decision they are trying to finalize about the UAT scope in the Risk: Calculation Discrepancy phase. The history contains two decisions in this phase from users with the same role, the validation scope stated by User\_13 and the release timing stated by User\_12. The BM25-based baseline ranks the record of User\_12 higher because of its lexical overlap with the question and answers with that decision, although the question concerns User\_13's own decision. \textsc{Org-Agent} records the asking user's identity as a node attribute and uses it to specify conditional filtering with \texttt{author = User\_13} and the phase name, so only records from User\_13 in this phase enter the candidate set, and the answer is the validation-scope decision. This case shows that \textsc{Org-Agent} accounts for the asking user's identity and information attribution when selecting records from the relevant project phase.

\obs{Dependency-aware scheduling resolves a required participant's availability before wider coordination.} In Figure~\ref{fig:case_study}~(b), the task is to schedule a meeting for multiple users whose attendance status is not given to the agent in advance. The vanilla baseline pursues Wednesday at 15:00, which most participants support, and finalizes it although Eve has repeatedly stated that she cannot attend, so the schedule excludes a required participant and fails the task. \textsc{Org-Agent} first asks the participants and learns that Eve and David are required, and that David lists Friday at 14:00 as a backup slot. It then places the interaction with David before those with the other participants to clarify whether he can use Friday at 14:00, since his confirmation is a prerequisite of proposing that slot. After David replies that he can attend if necessary, the TDG is updated to reflect the satisfied prerequisite, the remaining confirmation requests can proceed in parallel, and the meeting is finalized for Friday at 14:00 with both required participants. This case shows that \textsc{Org-Agent} obtains prerequisite confirmations before the interactions that depend on them.

\end{document}